\documentclass[runningheads]{llncs}
\usepackage{graphicx}

\usepackage{tikz}
\usepackage{comment} 
\usepackage{amsmath,amssymb} 
\usepackage{color}

\usepackage{multirow}
\usepackage{colortbl}
\usepackage{verbatim}
\usepackage{bm}

\begin{document}
\pagestyle{headings}
\mainmatter
\def\ECCVSubNumber{100}  

\title{A Boundary Based Out-of-Distribution Classifier for Generalized Zero-Shot Learning} 

\titlerunning{A Boundary Based OOD Classifier for Generalized Zero-Shot Learning}
%
\author{Xingyu Chen\inst{1}\orcidID{0000-0002-5226-963X} \and
Xuguang Lan\inst{1}\orcidID{0000-0002-3422-944X} \and
Fuchun Sun\inst{2}\orcidID{0000-0003-3546-6305} \and Nanning Zheng \inst{1}\orcidID{0000-0003-1608-8257}}
\authorrunning{X. Chen et al.}
%
\institute{Xi’an Jiaotong University, Xi'an, China,  \and
Tsinghua University, Beijing, China \\
\email{xingyuchen1990@gmail.com}, \email{\{xglan,nnzheng\}@mail.xjtu.edu.cn}, \email{fcsun@tsinghua.edu.cn}}
\maketitle

\begin{abstract}
Generalized Zero-Shot Learning (GZSL) is a challenging topic that has promising prospects in many realistic scenarios. Using a gating mechanism that discriminates the unseen samples from the seen samples can decompose the GZSL problem to a conventional Zero-Shot Learning (ZSL) problem and a supervised classification problem. However, training the gate is usually challenging due to the lack of data in the unseen domain. To resolve this problem, in this paper, we propose a boundary based Out-of-Distribution (OOD) classifier which classifies the unseen and seen domains by only using seen samples for training. First, we learn a shared latent space on a unit hyper-sphere where the latent distributions of visual features and semantic attributes are aligned class-wisely. Then we find the boundary and the center of the manifold for each class. By leveraging the class centers and boundaries, the unseen samples can be separated from the seen samples. After that, we use two experts to classify the seen and unseen samples separately. We extensively validate our approach on five popular benchmark datasets including AWA1, AWA2, CUB, FLO and SUN. The experimental results demonstrate the advantages of our approach over state-of-the-art methods.
\end{abstract}

\section{Introduction}
Zero-Shot Learning (ZSL) is an important topic in the computer vision community which has been widely adopted to solve challenges in real-world tasks. In the conventional setting, ZSL aims at recognizing the instances drawn from the unseen domain, for which the training data are lacked and only the semantic auxiliary information is available. However, in real-world scenarios, the instances are drawn from either unseen or seen domains, which is a more challenging task called Generalized Zero-Shot Learning (GZSL).
\begin{figure*}[t]
\begin{center}
    \includegraphics[width=0.7\linewidth]{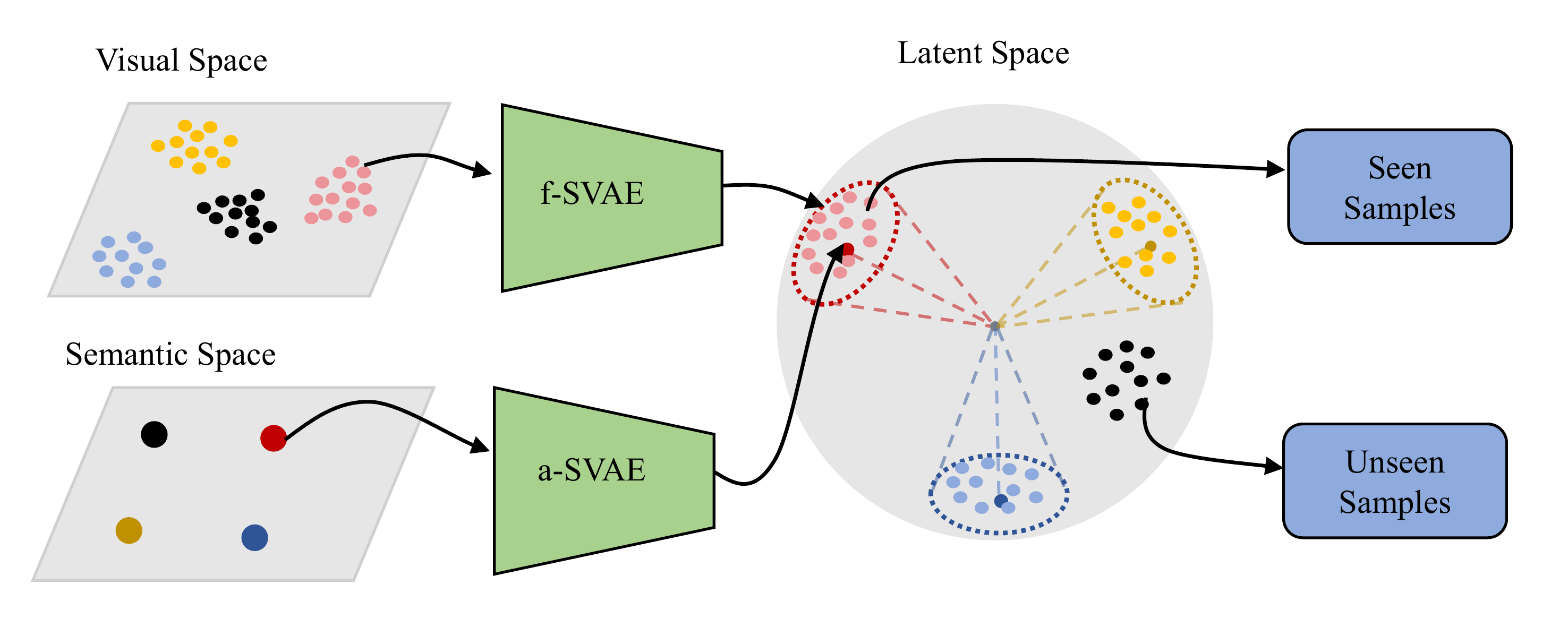}
\end{center}
   \caption{The boundary based OOD classifier learns a bounded manifold for each seen class on a unit hyper-sphere (latent space). By using the manifold boundaries (dotted circles) and the centers (dark-colored dots), the unseen samples (black dots) can be separated from the seen samples (colored dots). }
\label{fig:figure1}
\end{figure*}

Previous GZSL algorithms can be grouped into three lines: (1) Embedding methods \cite{akata2015evaluation,akata2016label,annadani2018preserving,kodirov2017semantic,romera2015embarrassingly,liu2018zero,tsai2017learning,socher2013zero,changpinyo2016synthesized,frome2013devise,zhang2019co} which aim at learning embeddings that unify the visual features and semantic attributes for similarity measurement. However, due to the bias problem \cite{zhang2019co}, the projected feature anchors of unseen classes may be distributed too near to that of seen classes in the embedding space. Consequently, the unseen samples are easily classified into nearby seen classes. (2) Generative methods \cite{mishra2018generative,chen2018zero,xian2018feature,felix2018multi,kumar2018generalized,schonfeld2019generalized} which focus on generating synthetic features for unseen classes by using generative models such as GAN \cite{goodfellow2014generative} or VAE \cite{kingma2013auto}. By leveraging the synthetic data, the GZSL problem can be converted to a supervised problem. Although the generative methods substantially improve the GZSL performance, they are still bothered by the feature confusion problem \cite{li2019alleviating}. Specifically, the synthetic unseen features may be tangled with the seen features. As a result, a classifier will be confused by the features which have strong similarities but different labels. An intuitive phenomenon is that previous methods usually make trade-offs between the accuracy of seen classes and unseen classes to get higher Harmonic Mean values. (3) Gating methods \cite{atzmon2019adaptive,socher2013zero} which usually incorporates a gating mechanism with two experts to handle the unseen and seen domains separately. Ideally, if the binary classifier is reliable enough, the GZSL can be decomposed to a ZSL problem and a supervised classification problem, which does not suffer from the feature confusion or bias problem. Unfortunately, it is usually difficult to learn such a classifier because unseen samples are not available during training. 

To resolve the main challenge in the gating methods, we propose a boundary based Out-of-Distribution (OOD) classifier for GZSL in this paper. As illustrated in Fig. \ref{fig:figure1}, the key idea of our approach is to learn a bounded manifold for each seen class in the latent space. A datum that can be projected into the bounded manifold will be regarded as a seen sample. Otherwise, we believe it is an unseen sample. In this way, we can easily separate unseen classes from seen classes even we do not use any unseen samples for training.

To learn a bounded manifold for each seen class, the proposed OOD classifier learns a shared latent space for both visual features and the semantic attributes. In the latent space, the distributions of visual features and semantic attributes are aligned class-wisely. Different from previous latent distribution aligning approach \cite{schonfeld2019generalized}, we build the latent space on a unit hyper-sphere by using Hyper-Spherical Variational Auto-Encoders (SVAE) \cite{s-vae18}. Specifically, the approximated posterior of each visual feature is encouraged to be aligned with a von Mises-Fisher (vMF) distribution, where the mean direction and concentration are associated with the corresponding semantic attribute. Therefore, each class can be represented by a vMF distribution on the unit hyper-sphere, which is easy to find the manifold boundary. In addition, the mean direction predicted by semantic attribute can be regarded as the class center. By leveraging the boundary and the class center, we can determine if a datum is projected into the manifold. In this way, the unseen features can be separated from the seen features. After that, we apply two experts to classify the seen features and unseen features separately.

The proposed classifier can incorporate with any state-of-the-art ZSL method. The core idea is very straightforward and easy to implement. We evaluate our approach on five popular benchmark datasets, i.e. AWA1, AWA2, CUB, FLO and SUN for generalized zero-shot learning. The experimental results demonstrate the advantages of our approach over state-of-the-art methods.

\section{Related Work}

\subsubsection{Embedding Methods}
To solve GZSL, the embedding methods \cite{akata2016label,akata2015evaluation,annadani2018preserving,kodirov2017semantic,romera2015embarrassingly,liu2018zero,tsai2017learning,socher2013zero,changpinyo2016synthesized,frome2013devise,zhang2019co,zhang2015zero,long2017zero,xu2017matrix,zhang2018zero} usually learn a mapping to unify the visual features and semantic attributes for similarity measurement. For example, Zhang et al. \cite{zhang2015zero} embed features and attributes into a common space where each point denotes a mixture of seen class proportions. Other than introducing common space, Kodirov et al. \cite{kodirov2017semantic} propose a semantic auto-encoder which aims to embed visual feature vector into the semantic space while constrain the projection must be able to reconstruct the original visual feature. On the contrary, Long et al. \cite{long2017zero} learn embedding from semantic space into visual space. However, due to the bias problem, previous embedding methods usually misclassify the unseen classes into seen classes. To alleviate the bias problem, Zhang et al. \cite{zhang2019co} propose a co-representation network which adopts a single-layer cooperation module with parallel structure to learn a more uniform embedding space with better representation. 

\subsubsection{Generative Methods} 

The generative methods \cite{mishra2018generative,chen2018zero,xian2018feature,felix2018multi,kumar2018generalized,schonfeld2019generalized} treat GZSL as a case of missing data and try to generate synthetic samples of unseen classes from semantic information. By leveraging the synthetic data, the GZSL problem can be converted to a supervised classification problem. Therefore, These methods usually rely on generative models such as GAN \cite{goodfellow2014generative} and VAE \cite{kingma2013auto}. For example, Xian et al. \cite{xian2018feature} directly generate image features by pairing a conditional WGAN with a classification loss. Mishara et al \cite{mishra2018generative} utilize a VAE to generate image features conditional on the class embedding vector. Felix et al. \cite{felix2018multi} propose a multi-modal cycle-consistent GAN to improve the quality of the synthetic features. Compared to the embedding methods, the generative methods significantly  improve the GZSL performance. However, Li et al. \cite{li2019alleviating} find that the generative methods are bothered by the feature confusion problem. To alleviate this problem, they present a boundary loss which maximizes the decision boundary of seen categories and unseen ones while training the generative model.

\subsubsection{Gating Methods}
There are a few works using a gating based mechanism to separate the unseen samples from the seen samples for GZSL. The gate usually incorporates two experts to handle seen and unseen domains separately. For example, Socher et al. \cite{socher2013zero} propose a hard gating model to assign test samples to each expert. Only the selected expert is used for prediction, ignoring the other expert. Recently, Atzmon et al. \cite{atzmon2019adaptive} propose a soft gating model which makes soft decisions if a sample is from a seen class. The key to the soft gating is to pass information between three classifiers to improve each one’s accuracy. Different from the embedding methods and the generative methods, the gating methods do not suffer from the bias problem or the feature confusion problem. However, a key difficulty in gating methods is to train a binary classifier by only using seen samples. In this work, we propose a boundary based OOD classifier by only using seen samples for training. The proposed classifier is a hard gating model which could provide accurate classification results.

\section{Revisit Spherical Variational Auto-Encoders}
The training objective of a general variational auto-encoder
is to maximize $\log \int p_{\phi }(x,z)dz$, the log-likelihood of the observed data, where $x$ is the training data, $z$ is the latent variable and $p_{\phi }(x,z)$ is a parameterized model representing the joint distribution of $x$ and $z$. However, computing the marginal distribution over the latent variable $z$ is generally intractable. In practice, it is implemented to maximize the Evidence Lower Bound (ELBO).
\begin{equation}
    \log \int p_{\phi}(x,z)dz \geq \mathbb{E}_{q(z))}\left [ \log p_{\phi }(x|z) \right ] - KL(q(z)||p(z)),
\end{equation}
where $q(z)$ approximates the true posterior distribution and $p(z)$ is the prior distribution. $p_{\phi}(x|z)$ is to map a latent variable to a data point $x$ which is parameterized by a decoder network.  $KL(q(z)||p(z))$ is the Kullback-Leibler divergence which encourages $q(z)$ to match the prior distribution. The main difference for various variational auto-encoders is in the adopted distributions.

For SVAE \cite{s-vae18}, both of the prior and posterior distributions are based on von Mises-Fisher(vMF) distributions. A vMF distribution can be regarded as a Normal distribution on a hyper-sphere, which is defined as:
\begin{equation}
q(z|\mu, \kappa ) = C_{m}(\kappa)\exp (\kappa\mu^Tz)
\end{equation}
\begin{equation}
C_{m}(\kappa) = \frac{\kappa^m/2-1}{(2\pi)^{m/2}I_{m/2-1}(\kappa)}
\end{equation}
where $\mu\in \mathbb{R}^{m}, ||\mu||_{2}=1$ represents the direction on the sphere and $\kappa\in\mathbb{R}_{\geq0}$ represents the concentration around $\mu$. $C_{m}(\kappa)$ is the normalizing constant, $I_{v}$ is the modified Bessel function of the first kind at order $v$.

Theoretically, $q(z)$ should be optimized over all data points, which is not tractable for large dataset. Therefore it uses $q_{\theta}(z|x)=q(z|\mu(x),\kappa(x))$ which is parameterized by an encoder network to do stochastic gradient descent over the dataset. The final training objective is defined as:
\begin{equation}
    L_{\rm{SVAE}}(\theta,\phi;x) = \mathbb{E}_{q_{\theta}(z|x)}  \left [ \log p_{\phi }(x|z) \right ]-KL(q_{\theta}(z|x)||p(z)).
\end{equation}

\section{Proposed Approach}

\subsection{Problem Formulation}
We first introduce the definitions of OOD classification and GZSL. We are given a set of training samples of seen classes $\mathcal{S}=\left \{(x,y,a)|x\in \mathcal{X},y\in \mathcal{Y}_s,a\in \mathcal{A}_s \right \}$ where $x$ represents the feature of an image extracted by a CNN, $y$ represents the class label in $\mathcal{Y}_s=\left \{ y^{1}_s,y^2_s,...,y^N_s \right \}$ consisting of $N$ seen classes and $a$ represents corresponding class-level semantic attribute which is usually hand-annotated or a Word2Vec feature \cite{mikolov2013distributed}. We are also given a set $\mathcal{U}=\left \{(y,a)|y\in \mathcal{Y}_u,a\in \mathcal{A}_u \right \}$  of unseen classes  $\mathcal{Y}_{u}=\left \{y^1_u,y^2_u,...,y^M_u \right \}$. The zero shot recognition states that $\mathcal{Y}_s \cap \mathcal{Y}_u= \varnothing$. Given $\mathcal{S}$ and $\mathcal{U}$, the OOD classifier aims at learning a binary classifier $f_{OOD}: \mathcal{X} \rightarrow \{0,1 \}$ that distinguishes if a datum belongs to $\mathcal{S}$ or $\mathcal{U}$. The task of GZSL aims at learning a classifier $f_{gzsl}: \mathcal{X} \rightarrow \mathcal{Y}_s \cup \mathcal{Y}_u $.

\subsection{Boundary Based Out-of-Distribution Classifier}
\begin{figure*}[t]
\begin{center}
    \includegraphics[width=0.85\linewidth]{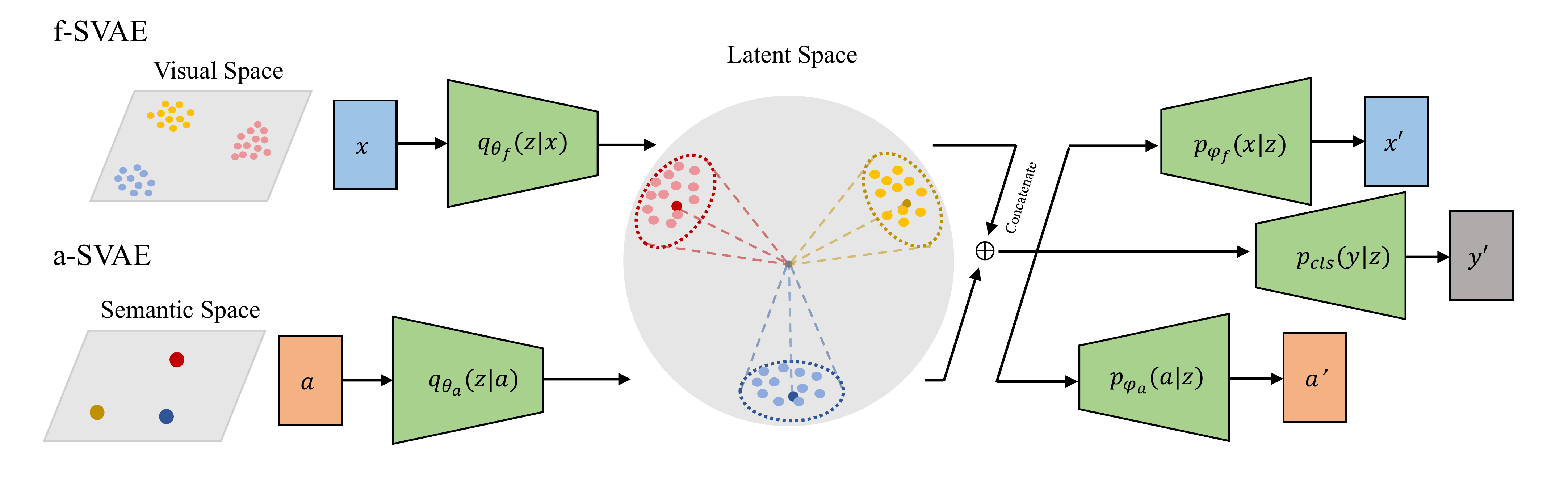}
\end{center}
   \caption{Our model consists of two SVAEs, one for visual features and another for semantic attributes. By combining the objective functions of the two SVAEs with a cross-reconstruction loss and a classification loss, we train our model to align the latent distributions of visual features and semantic attributes class-wisely. In this way, each class can be represented by a vMF distribution whose boundary is easy to find. }
\label{fig:framework}
\end{figure*}

The proposed OOD classifier aims to classify the unseen and seen domains by only using seen samples for training. The core idea of our approach is quite straightforward. First, we build a latent space for visual features and semantic attributes. Then we learn a bounded manifold for each seen class. Next we find the boundaries of the learned manifolds. By leveraging the boundaries, we can determine if a test sample is projected into the manifolds. For the samples which can be projected into the manifolds, we believe they belong to the seen domain and assign them to a seen expert. Otherwise, we assign them to an unseen expert.

\subsubsection{Build the Latent Space on a Unit Hyper-sphere}
Different from previous works, we build the latent space on a unit hyper-sphere by using hyper-spherical variational auto-encoders. In the latent space, each class is approximately represented by a vMF distribution of which the mean direction can be regarded as the class center. Using the spherical representation has two advantages. First, we can naturally use cosine similarity as the distance metric since all latent variables and class centers are located on the unit hyper-sphere. Second, for each seen class, it is easy to find the manifold boundary. Specifically, we can find a threshold based on the cosine similarities between the latent variables and the class center. According to the class center and the corresponding boundary, we can determine if a visual feature is projected into the manifold.

\subsubsection{Learn A Bounded Manifold for Each Class}

To learn a bounded manifold for each class, inspired by \cite{schonfeld2019generalized}, we encourage the latent distributions of visual features and the corresponding semantic attribute to be aligned with each other in the latent space. As illustrated in Fig. \ref{fig:framework}, our model consists of two SVAEs correspond to two data modalities, one for visual features and another for semantic attributes, denoted as f-SVAE and a-SVAE, respectively. Given an attribute $a \in \mathcal{A}_s$, the encoder of a-SVAE predicts a vMF distribution $q_{\theta_a}(z|a)=q(z|\mu(a),\kappa(a))$. Meanwhile, given the corresponding visual feature $x$, the encoder of f-SVAE predicts a vMF distribution $q_{\theta_f}(z|x)=q(z|\mu(x),\kappa(x))$. Each SVAE regards the distribution predicted by another SVAE as the prior distribution. Therefore, we can align the two distributions by optimizing the objective functions of f-SVAE and a-SVAE simultaneously. We further adopt a cross-reconstruction loss and a classification loss to ensure the latent representations capture the modality invariant information while preserving discrimination. Therefore, the training objective consists four parts.

\textit{f-SVAE}: For the f-SVAE, we expect to maximize the log-likelihood and minimize the discrepancy between the approximated posterior $q_{\theta_f}(z|x)$  and the prior distribution $q_{\theta_a}(z|a)$. Therefore, the training objective is defined as:
\begin{equation}
    L_{f-SVAE}= \mathbb{E}_{p(x,a)}[ \mathbb{E}_{q_{\theta_f}(z|x)} [\log{p_{\phi_{f}}(x|z)}]-  \lambda_{f} D_{z} (q_{\theta_{f}}(z|x)\parallel q_{\theta_a}(z|a))],
\end{equation}
where $\mathbb{E}_{q_{\theta_f}(z|x)} [\log{p_{\phi_{f}}(x|z)}]$ represents the expectation of log-likelihood over latent variable $z$. In practice, we use the negative reconstruction error of visual feature $x$ instead. $p_{\phi_{f}}(x|z)$ is the decoder network of f-SVAE. $D_{z} (q_{\theta_{f}}(z|x)\parallel q_{\theta_a}(z|a))$ represents the discrepancy between the two vMF distributions. $\lambda_f$ is a hyper-parameter to weight the discrepancy term. It worth noting that $D_{z}(q_{\theta_{f}}(z|x)\parallel q_{\theta_a}(z|a))$ is the Earth Mover's Distance (EMD) between the two distributions which is defined as:
\begin{equation}
    D_{z} (q_{\theta_{f}}(z|x)\parallel q_{\theta_a}(z|a))=\inf_{\Omega\in \prod(q_{\theta_{f}}, q_{\theta_a})} \mathbb{E}_{(z_1,z_2) \sim \Omega}[\parallel z_1-z_2 \parallel].
\end{equation}
The reason we use EMD instead of the KL-divergence is that the KL-divergence may fail when the support regions of the two distributions  $q_{\theta_{f}}(z|x)$ and $q_{\theta_{a}}(z|a)$ do not completely coincide. To calculate the EMD, we utilize the Sinkhorn iteration algorithm in \cite{cuturi2013sinkhorn}.  

\textit{a-SVAE}: Similarly, for the a-SVAE, $q_{\theta_f}(z|x)$ is regarded as the prior distribution. The objective function is defined as:
\begin{equation}
    L_{a-SVAE}= \mathbb{E}_{p(x,a)}[ \mathbb{E}_{q_{\theta_a}(z|a)} [\log{p_{\phi_{a}}(a|z)}]- \lambda_a D_{z} ( q_{\theta_a}(z|a) \parallel q_{\theta_{f}}(z|x))],
\end{equation}
where $\mathbb{E}_{q_{\theta_a}(z|a)} [\log{p_{\phi_{a}}(a|z)}]$ represents the negative reconstruction error of semantic attribute $a$. $D_{z} ( q_{\theta_a}(z|a) \parallel q_{\theta_{f}}(z|x))$ is the discrepancy between the two vMF distributions.
As EMD is symmetrical, $D_{z} ( q_{\theta_a}(z|a) \parallel q_{\theta_{f}}(z|x))$ equals to  $ D_{z} (q_{\theta_{f}}(z|x)\parallel q_{\theta_a}(z|a))$, weighted by hyper-parameter $\lambda_a$.

\textit{Cross-reconstruction Loss}: Since we learn a shared latent space for the two different modalities, the latent representations should capture the modality invariant information. For this purpose, we also adopt a cross-reconstruction regularizer:

\begin{equation}
    L_{cr}= \mathbb{E}_{p(x,a)}[ \mathbb{E}_{q_{\theta_a}(z|a)} [\log{p_{\phi_{f}}(x|z)}] +  \mathbb{E}_{q_{\theta_f}(z|x)} [\log{p_{\phi_{a}}(a|z)}]],
\end{equation}
where $\mathbb{E}_{q_{\theta_a}(z|a)} [\log{p_{\phi_{f}}(x|z)}]$ and $\mathbb{E}_{q_{\theta_f}(z|x)} [\log{p_{\phi_{a}}(a|z)}]$ also represent negative reconstruction errors.

\textit{Classification Loss}: To make the latent variables more discriminate, we introduce the following classification loss:
\begin{equation}
    L_{cls}= \mathbb{E}_{p(x,y,a)}[ \mathbb{E}_{q_{\theta_a}(z|a)} [\log{p_{\phi_{cls}}(y|z)}] +  \mathbb{E}_{q_{\theta_f}(z|x)} [\log{p_{\phi_{cls}}(y|z)}]],
\end{equation}
where $\phi_{cls}$ represents  the  parameters  of a linear softmax classifier. Although the classification loss may hurt the inter-class association between seen and unseen classes, it also reduces the risk for unseen features being projected into the manifolds of seen classes, which benefits to the binary classification. The reason is that our OOD classifier only cares about separating unseen features from the seen features, but not cares about which class the unseen features belong to.

\textit{Overall Objective}: Finally, we train our model by maximizing the following objective:
\begin{equation}
    L_{overall}=L_{f-SVAE} + L_{a-SVAE}+ \alpha L_{cr} + \beta L_{cls},
\end{equation}
where $\alpha$, $\beta$ are the hyper-parameters used to weight the two terms.

\subsubsection{Find the Boundaries for OOD Classification}
When the proposed model is trained to convergence, the visual features and the semantic attributes are aligned class-wisely in the latent space. Each class is represented by a vMF distribution. Therefore, the manifold of each class can be approximately represented by a circle on the unit hyper-sphere. By leveraging the center and the boundary, we can determine whether a latent variable locates in the manifold.

For class $y^i \in \mathcal{Y}_s $, the class center can be found by using its semantic attribute.  Given $a^i \in \mathcal{A}_s$, a-SVAE predicts a vMF distribution $q(z|\mu(a^i),\kappa(a^i))$ of which $\mu(a^i)$ is regarded as the class center. 


There could be many ways to find the boundaries. For example, the simplest method is to search a threshold by cross validation to represent boundaries. Another simple strategy is to leveraging the statistics of training samples. For $i$-th seen class, we first encode all training samples to latent variables. After that we calculate the cosine similarity $S(\bm{z}^i, \mu(\bm{a}^i))$ between each latent variable $\bm{z}^i$ and the class center $\bm{\mu}(\bm{a}^i)$. Then we search a threshold $\eta^i$ which is smaller than $\gamma \in (0,100\%)$ and larger than $1-\gamma$ of the cosine similarities. We adopt $\eta^i$ to represent the boundary of class $i$. Here, $\gamma$ can be viewed as the OOD classification accuracy on training samples. Given a $\gamma$, we can find the corresponding threshold $\eta$ for each seen class.

Given a test sample $\bm{x}$ which may come from a seen class or an unseen class, we first encode it to latent variable $\bm{z}$. Then we compute the cosine similarities between it to all seen class centers and find the maximum. So we can get the closest manifold boundary $\eta^*$. By leveraging the threshold $\eta^*$, we determine the test sample belongs to unseen class or seen class using Eq.(11),
\begin{equation}
y^{OOD}=\left\{\begin{matrix}
        0, \quad\text{if} \quad  \max \{S(\bm{z},\bm{\mu}(\bm{a}^i))|\forall \bm{a}^i \in \mathcal{A}_s\}< \eta^*\\ 
        1, \quad\text{if} \quad \max \{S(\bm{z},\bm{\mu}(\bm{a}^i))|\forall \bm{a}^i \in \mathcal{A}_s\}\geq \eta^*\\
\end{matrix}\right.
\end{equation}
where 0 stands for unseen class and 1 for seen class. 
 
\subsubsection{Generalized Zero-Shot Classification}
For the GZSL task, we incorporate the proposed OOD classifier with two domain experts. Given a test sample, the OOD classifier determines if it comes from a seen class. Then, according to the predicted label, the test sample is assigned to a seen expert or an unseen expert for classification.

\subsection{Implementation Details}
\subsubsection{OOD Classifier}
For the f-SVAE, we use two 2-layer Fully Connected (FC) network for the encoder and decoder networks. The first FC layer in the encoder has 4096 neurons with ReLU followed. The output is then fed to two FC layers to produce the mean direction and the concentration for the reparameterize trick. The mean direction layer has $128$ neurons and the concentration layer only has $1$ neuron. The output of mean direction layer is normalized by its norm such that it lies on the unit hyper-sphere. The concentration layer is followed by a Softplus activation to ensure its output larger than $0$. The decoder consists of two FC layers. The first layer has 4096 neurons with ReLU followed. The number of neurons in the second layer equals to the dimension of the input features.

The structure of a-SVAE is similar to f-SVAE except for the input dimension and the neuron number of the last FC layer equal to the dimension of the semantic attributes. We use a Linear Softmax classifier which takes the latent variables as input for calculating the classification loss. The structure is same as in \cite{xian2018feature}.

We train our model by the Adam optimizer with learning rate $0.001$. The batch size is set to $128$. The hyper-parameter $\lambda_f$, $\lambda_a$, $\alpha$, $\beta$ are set to $1.0$, respectively.

\subsubsection{Unseen and Seen Experts}
For the unseen samples, we use f-CLSWGAN \cite{xian2018feature} and f-VAEGAN \cite{xian2019f}. For the seen samples, we directly combine the encoder of f-SVAE and the linear softmax classifier for classification.

\section{Experiments}
The proposed approach is evaluated on five benchmark datasets, where plenty of recent state-of-the-art methods are compared. Moreover, the features and settings used in experiments follow the paper \cite{xian2018zero} for fair comparison.

\subsection{Datasets, Evaluation and Baselines}

\subsubsection{Datasets} 
The five benchmark datasets include Animals With Attributes 1 (AWA1)\cite{2014attribute}, Animal With Attributes 2 (AWA2) \cite{xian2018zero}, Caltech-UCSD-Birds (CUB) \cite{wah2011caltech}, FLOWER (FLO) \cite{nilsback2008automated} and SUN attributes (SUN) \cite{patterson2014sun}. Specifically, AWA1 contains 30,475 images and 85 kinds of properties, where 40 out of 50 classes are obtained for training. In AWA2, 37,322 images in the same classes are re-collected because original images in AWA1 are not publicly available. CUB has 11,788 images from 200 different types of birds annotated with 312 properties, where 150 classes are seen and the others are unseen during training. FLO consists of 8,189 images which come from 102 flower categories, where 82/20 classes are used for training and testing. For this dataset, we use the same semantic descriptions provided by \cite{reed2016learning}. SUN has 14,340 images of 717 scenes annotated with 102 attributes, where 645 classes are regarded as seen classes and the rest are unseen classes. 

\subsubsection{Evaluation}  For OOD classification, the in-distribution samples are regarded as the seen samples and the out-of-distribution samples are regarded as unseen samples. The True-Positive-Rate (\textbf{TPR}) indicates the classification accuracy of seen classes and the False-Positive-Rate (\textbf{FPR}) indicates the accuracy of unseen classes. We also measure the Area-Under-Curve (\textbf{AUC}) by sweeping over classification threshold.

For GZSL, the average of per-class precision (AP) is measured. The ``\textbf{ts}'' and ``\textbf{tr}'' denote the Average Precision (AP) of images from unseen and seen classes, respectively. ``\textbf{H}'' is the harmonic mean which is defined as: $H = 2* tr* ts / (tr+ts)$. The harmonic mean reflects the ability of method that recognizes seen and unseen images simultaneously.

\subsubsection{GZSL Baselines} We compare our approach with three lines of previous works in the experiments. (1) Embedding methods which focus on learning embeddings that unify the visual features and semantic attributes for similarity measurement. We include the recent competitive baselines: SJE \cite{akata2015evaluation}, ALE \cite{akata2016label}, PSR \cite{annadani2018preserving}, SAE \cite{kodirov2017semantic}, EZSL \cite{romera2015embarrassingly}, LESAE \cite{liu2018zero}, ReViSE \cite{tsai2017learning}, CMT \cite{socher2013zero}, SYNC \cite{changpinyo2016synthesized}, DeViSE \cite{frome2013devise} and CRnet \cite{zhang2019co}. (2) Generative methods which focus on generating synthetic features or images for unseen classes using GAN or VAE. We also compare our approach with the recent state-of-the-arts such as CVAE \cite{mishra2018generative}, SP-AEN \cite{chen2018zero}, f-CLSWGAN \cite{xian2018feature}, CADA-VAE \cite{schonfeld2019generalized}, cycle-(U)WGAN \cite{felix2018multi}, SE \cite{kumar2018generalized} and AFC-GAN \cite{li2019alleviating}, f-VAEGAN \cite{xian2019f}. (3) Gating methods which aim at learning a classifier to distinguish the unseen features from the seen features. We compare our approach with the recent state-of-the-art COSMO \cite{atzmon2019adaptive}.

\subsection{Out-of-Distribution Classification}

\begin{table*}[t]
    \scriptsize
	\renewcommand{\multirowsetup}{\centering}
	\renewcommand\arraystretch{1.0}
	\centering
	\caption{Comparison with various gating models on validation set. \textbf{AUC} denotes Area-Under-Curve when sweeping over detection threshold. \textbf{FPR} denotes False-Positive-Rate on the threshold that yields $95\%$ True Positive Rate for detecting in-distribution samples.}
	\centering
	
	\begin{tabular}{p{4.5cm}<{\centering}|
			p{0.75cm}<{\centering}p{0.75cm}<{\centering}|
			p{0.75cm}<{\centering}p{0.75cm}<{\centering}|
			p{0.75cm}<{\centering}p{0.75cm}<{\centering}}
			
		\hline
		&\multicolumn{2}{c|}{AWA1}&		\multicolumn{2}{c|}{CUB} & \multicolumn{2}{c}{SUN} \\

		\hline
		Method&		\textbf{AUC}&	\textbf{FPR} & 	\textbf{AUC}&	\textbf{FPR}& 	\textbf{AUC}&	\textbf{FPR} \\
		\hline
		 MAX-SOFTMAX-1 \cite{hendrycks17baseline} & 86.7 & 67.9 & 74.1 & 82.4  &  60.9 & 92.8\\
		 MAX-SOFTMAX-3 \cite{hendrycks17baseline}  & 88.6 & 56.8  & 73.4 & 79.6 & 61.0 & 92.3 \\
		 CB-GATING-3 (w/o$P^{ZS}$) \cite{atzmon2019adaptive} & 88.8 & 56.4 & 74.2 & 80.1&  61.0 & 92.2 \\
		 CB-GATING-1 \cite{atzmon2019adaptive} & 88.9 & 59.1 & 81.7 & 73.1  & 75.5 & 77.5 \\
		 CB-GATING-3\cite{atzmon2019adaptive} &  92.5  & 45.5  & 82.0 & 72.0  &  77.7 & 77.5 \\ 
		 \textbf{Ours}  & 89.1 & 55.7   &71.2 & 85.0   & 63.1 & 88.8\\
	
		\hline
	\end{tabular}
	\label{tab:binary_val}
\end{table*}

We first conduct OOD classification experiments on the five benchmark datasets. We compare the boundary based OOD classifier with two state-of-the-art gating-based methods: (1) MAX-SOFTMAX-1 and MAX-SOFTMAX-3 are baseline gating models of \cite{hendrycks17baseline}. (2) CB-GATING-3 (w/o$P^{ZS}$), CB-GATING-1 and CB-GATING-3 are confidence-based gating models in \cite{atzmon2019adaptive}. 
For a fair comparison, we use the same dataset splitting as in \cite{atzmon2019adaptive}.
Table \ref{tab:binary_val} shows the classification results of the proposed OOD classifier compared to the baseline methods. It worth noting that the FPR scores are reported on the threshold that yields $95\%$ TPR for detecting in-distribution samples.

On AWA1, we see that our approach achieves $89.1\%$ AUC score which is the second best performance. The CB-GATING-3 performs better because it implicitly uses the semantic information of unseen classes during training. On CUB and SUN, our approach only achieves $71.2\%$ and $63.1\%$ AUC results. We think there are two reasons for these results. First, the data distribution of test seen set and train set not are so similar. For example, on SUN, the seen expert we use could only achieve $48.4\%$ top-1 accuracy. Second, under this data split, there are only $2938$ training samples for CUB and $6700$ for SUN, which is insufficient to learn accurate manifold boundaries.

\begin{figure*}[t]
\begin{center}
    \includegraphics[width=0.4\linewidth]{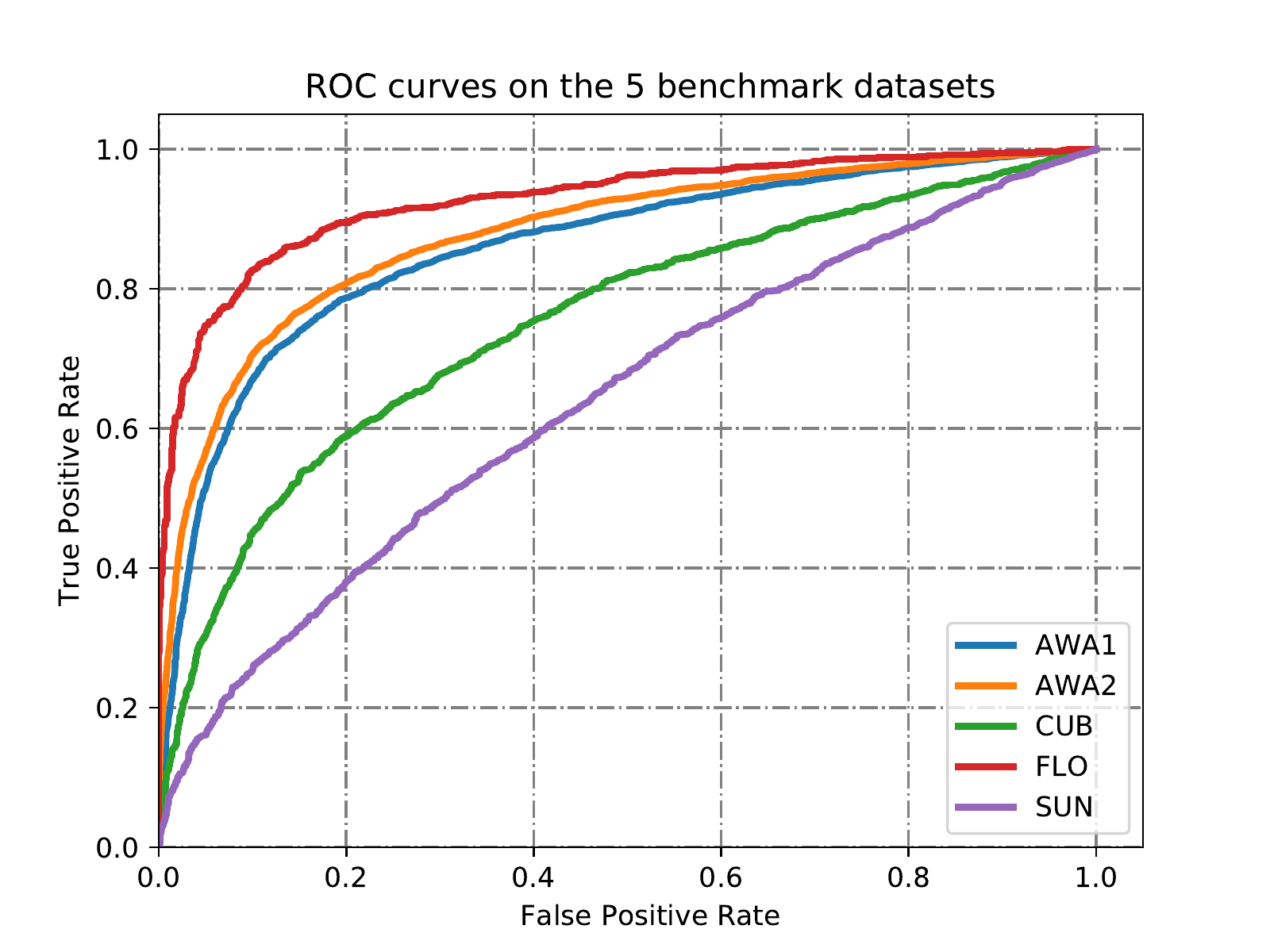}
\end{center}
   \caption{The ROC curves on the five benchmark datasets. }
\label{fig:roc}
\end{figure*}

We also present the OOD classification results on the test sets of the five benchmark datasets. The ROC curves are shown in Fig. \ref{fig:roc}, where the AUC score is $85.9\%$ on AWA1, $87.8\%$ on AWA2, $75.9\%$ on CUB, $92.7\%$ on FLO and $64.0\%$ on SUN. 

\subsection{Comparison with State-of-the-Arts}

\begin{table*}[t]
    \scriptsize
	\renewcommand{\multirowsetup}{\centering}
	\renewcommand\arraystretch{1.0}
	\centering
	\caption{Generalized Zero-Shot Learning results on AWA1, AWA2, CUB, FLO, and SUN. We measure the AP of Top-1 accuracy in \%. The best results are highlighted with bold numbers.}
	\centering
	\resizebox{0.95\linewidth}{!}{
	\begin{tabular}{p{3.5cm}<{\centering}|
			p{0.6cm}<{\centering}p{0.6cm}<{\centering}p{0.55cm}<{\centering}|
			p{0.6cm}<{\centering}p{0.6cm}<{\centering}p{0.55cm}<{\centering}|
			p{0.6cm}<{\centering}p{0.6cm}<{\centering}p{0.55cm}<{\centering}|
			p{0.6cm}<{\centering}p{0.6cm}<{\centering}p{0.55cm}<{\centering}|
			p{0.6cm}<{\centering}p{0.6cm}<{\centering}p{0.55cm}<{\centering}}
			
		\hline
		&\multicolumn{3}{c|}{AWA1}&	\multicolumn{3}{c|}{AWA2}&	\multicolumn{3}{c|}{CUB} & \multicolumn{3}{c|}{FLO} & \multicolumn{3}{c}{SUN}\\

		\hline
		Method&	\textbf{ts}	&\textbf{tr}&	\textbf{H}&	\textbf{ts}&	\textbf{tr}&	\textbf{H}&	\textbf{ts}&	\textbf{tr}&	\textbf{H} & \textbf{ts}&	\textbf{tr}&	\textbf{H}& \textbf{ts}&	\textbf{tr}&	\textbf{H}\\
		\hline
		SJE \cite{akata2015evaluation} & 11.3 & 74.6 & 19.6 & 8.0 & 73.9 & 14.4 & 23.5 & 59.2 & 33.6 & 13.9 & 47.6 & 21.5 & 14.7 & 30.5 & 19.8\\ 
		ALE \cite{akata2016label}  & 16.8 & 76.1 & 27.5 & 14.0 & 81.8 & 23.9 & 23.7 & 62.8 & 34.4 & 13.3 & 61.6 & 21.9 & 21.8 & 33.1 & 26.3\\
		PSR \cite{annadani2018preserving} \ & - & - & - & 20.7 & 73.8 & 32.3 & 24.6 & 54.3 & 33.9 & - & - & - & 20.8 & 37.2 & 26.7\\
		SAE \cite{kodirov2017semantic} & 16.7 & 82.5 & 27.8 & 8.0 & 73.9 & 14.4 & 18.8 & 58.5 & 29.0 & - & - & - & 8.8 & 18.0 & 11.8\\ 
		ESZSL \cite{romera2015embarrassingly}  & 6.6 & 75.6 & 12.1 & 5.9 & 77.8 & 11.0 & 12.6 & 63.8 & 21.0 & 11.4 & 56.8 & 19.0 & 11.0 & 27.9 & 15.8\\
		LESAE \cite{liu2018zero} & 19.1 & 70.2 & 30.0 & 21.8 & 70.6 & 33.3 & 24.3 & 53.0 & 33.3 & - & - & - & 21.9 & 34.7 & 26.9\\
		ReViSE \cite{tsai2017learning} & 46.1 & 37.1 & 41.1 & 46.4 & 39.7 & 42.8 & 37.6 & 28.3 & 32.3  & -& -& - & 24.3 & 20.1 & 22.0\\
		CMT \cite{socher2013zero}  & 0.9 & \textbf{87.6} & 1.8 & 0.5 & 90.0 & 1.0 & 7.2 & 49.8 & 12.6  & - & - & - & 8.1 & 21.8 & 11.8 \\
        SYNC \cite{changpinyo2016synthesized} & 8.9 & 87.3 & 16.2 & 10.0 & 90.5 & 18.0 & 11.5 & \textbf{70.9} & 19.8 & - & - & - & 7.9 & \textbf{43.3} & 13.4 \\
        DeViSE \cite{frome2013devise} & 13.4 & 68.7 & 22.4 & 17.1 & 74.7 & 27.8 & 23.8 & 53.0 & 32.8 & 9.9 & 44.2 & 16.2 & 16.9 & 27.4 & 20.9\\
        ALS \cite{li2020joint} & 53.8 & 56.0 & 54.9 & - & - & - & 43.1 & 51.6 & 46.9 & - & - & - & 41.5 & 31.9 & 36.1 \\
        CRnet \cite{zhang2019co} & 58.1 & 74.7 & 65.4 & 52.6 & 78.8 & 63.1 & 45.5 & 56.8 & 50.5 & - & - & - & 34.1 & 36.5 & 35.3\\
        \hline
        CVAE \cite{mishra2018generative}  & - & - & 47.2 & - & - & 51.2 & - & - & 34.5  & - & - & - & - & - & 26.7\\
        SP-AEN \cite{chen2018zero}  & - & - & - & 23.3 & \textbf{90.9} & 37.1 & 34.7 & 70.6 & 46.6 & - & - & - & 24.9 & 38.6 & 30.3\\
        f-CLSWGAN \cite{xian2018feature} & 57.9 & 61.4 & 59.6 & 52.1 & 68.9 & 59.4 & 43.7 & 57.7 & 49.7 & 59.0 & 73.8 & 65.6 & 42.6 & 36.6 & 39.4 \\
        cycle-(U)WGAN \cite{felix2018multi} & 59.6 & 63.4 & 59.8 & - & - & - & 47.9 & 59.3 & 53.0 & \textbf{61.6} & 69.2 & 65.2 &  47.2 & 33.8 & 39.4  \\
        SE \cite{kumar2018generalized}  & 56.3 & 67.8 & 61.5 & \textbf{58.3} & 68.1 & 62.8 & 41.5 & 53.3 & 46.7 & - & - & - & 40.9 & 30.5 & 34.9\\
        
        CADA-VAE \cite{schonfeld2019generalized} & 57.3 & 72.8 & \textbf{64.1} & 55.8 & 75.0 & 63.9 & 51.6 & 53.5 &52.4 & - &- &- & 47.2 & 35.7 & 40.6\\
        AFC-GAN \cite{li2019alleviating} & - & - & -& 58.2 & 66.8 & 62.2 & \textbf{53.5} & 59.7 & \textbf{56.4} & 60.2 & 80.0 & 68.7 & \textbf{49.1} & 36.1 & \textbf{41.6}\\
        f-VAEGAN \cite{xian2019f}  & - & - & -& 57.6 & 70.6 & 63.5 & 48.4 & 60.1 & 53.6 & 56.8 & 74.9 & 64.6 & 45.1 & 38.0 & 41.3\\
        
        \hline
        COSMO+f-CLSWGAN \cite{atzmon2019adaptive}  & \textbf{64.8} & 51.7 & 57.5 & - & - & - & 41.0 & 60.5 & 48.9 & 59.6 & 81.4 & 68.8 & 35.3 & 40.2 & 37.6\\
        COSMO+LAGO \cite{atzmon2019adaptive} & 52.8 & 80.0 & 63.6 & - & - & - & 44.4 & 57.8 & 50.2 & - & - & - & 44.9 & 37.7 & 41.0\\
        \hline
		
		\textbf{Ours}+f-CLSWGAN  & 51.9 & 72.7 & 60.6 & 53.4 & 75.9 & 62.7 & 46.8 & 50.2 & 48.4 & 59.4 & 81.2 & 68.6 &  37.6 & 33.9 & 35.6 \\
		\textbf{Ours}+f-VAEGAN  & 54.7 & 72.7 & 62.4 & 55.6 & 75.9 & \textbf{64.2} & 49.5 & 50.2 & 49.8 & 60.0 & 81.2 & \textbf{69.0} &  40.7 & 33.9 & 37.0 \\
	
		\hline
	\end{tabular}}
	\label{tab:gzsl}
\end{table*}

We further evaluate our approach on the five benchmark datasets under the GZSL setting. We report the top-1 accuracy and harmonic mean of each method in Table \ref{tab:gzsl} where ``-'' indicates that the result is not reported.

First, we see that most of the embedding methods suffer from the bias problem. For example, the ts values of baseline methods \cite{akata2015evaluation,akata2016label,annadani2018preserving,kodirov2017semantic,romera2015embarrassingly,liu2018zero,tsai2017learning,socher2013zero,changpinyo2016synthesized,frome2013devise} are much lower than the tr values, which leads to poor harmonic results. Compared to the embedding methods, the generative methods \cite{mishra2018generative,chen2018zero,xian2018feature,felix2018multi,kumar2018generalized,schonfeld2019generalized} show much higher harmonic mean results. However, due to the feature confusion problem, these methods have to make trade-offs between ts and tr values to get higher harmonic mean results. For example, on AWA2, without considering the GZSL task, we could get $92.8\%$ top-1 classification accuracy on seen classes. However, when adding the synthetic unseen features for training, the tr values of f-CLSWGAN, cycle-(U)WGAN, SE, CADA-VAE and AFC-GAN clearly decrease. Using a gating mechanism to distinguish seen and unseen samples is also a reliable way to solve GZSL problem. We see that the gating based method COSMO achieves competitive results compared to the generative methods.

Second, our approach achieves superior performance compared to the previous methods on AWA2 and FLO, achieves competitive results on AWA1, CUB, and SUN. Specifically, our approach achieves $62.4\%$ harmonic mean on AWA1, $64.2\%$ on AWA2, $49.8\%$ on CUB, $69.0\%$ on FLO, and $37.0\%$ on SUN. We can observe that the results of Ours+f-CLSWGAN are close to COSMO+f-CLSWGAN. Noting that COSMO relies on both seen and unseen experts during the training stage, which means the semantic information of unseen classes is utilized. Our approach only use the semantic information of seen classes for training. During testing, our approach uses the semantic attributes of seen classes. 

Third, in our experiment, we combine our OOD classifiers with the unseen experts. Take f-VAEGAN for an example, compared to f-VAEGAN, the harmonic mean results of Ours+f-VAEGAN are improved by $0.7\%$ on AWA2, $4.4\%$ on FLO. However, on CUB and SUN, we observe that our approach achieves lower harmonic mean results. For AWA1, AWA2 and FLO, the data distributions of test seen set and train set are similar. The test samples of seen classes could be easily projected into the learned manifolds in the latent space. Therefore, using the manifold boundaries to separate the seen and unseen samples is quite effective. However, on CUB and SUN, the data distributions of test seen set and train set are very different. In this situation, the manifold boundaries cannot separate the seen and unseen samples very well.   

Obviously, the GZSL performance of our approach mainly depends on the OOD classifier and the ZSL classifier. When the test set and train set of seen classes have a similar data distribution, our classifier is reliable enough to separate the unseen features from the seen features. At this time, the GZSL problem can be substantially simplified. In practice, we can replace the ZSL classifier by any state-of-the-art models. Consequently, the Harmonic mean of our approach could be further improved by using more powerful ZSL models.

\subsection{Model Analysis}

\subsubsection{Latent Space Visualization}
\begin{figure*}[t]
\begin{center}
    \includegraphics[width=1.0\linewidth]{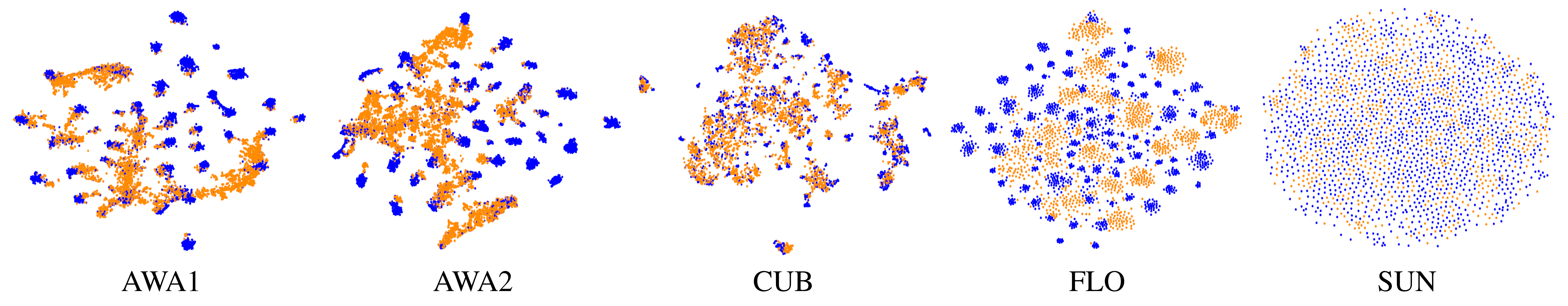}
\end{center}
   \caption{The t-SNE visualization results for the learned latent space on the test sets of AWA1, AWA2, CUB, FLO and SUN. The blue dots represent the variables encoded from seen classes. The orange dots represent the variables encoded from unseen classes. }
\label{fig:latent}
\end{figure*}

To demonstrate the learned latent space, we visualize the latent variables of seen features and unseen features to the 2D-plane by using t-SNE. The visualization results of five datasets are shown in Fig. \ref{fig:latent} where the blue dots represent the seen variables and the orange dots represent the unseen variables. It can be seen that the features in each seen class are clustered together in the latent space so that they can be easily classified. The unseen features are encoded to the latent variables chaotically scattered across the latent space. We see that most of the unseen variables locate out of the manifolds of seen classes. Although the inter-class association between seen and unseen classes is broken, the unseen variables can be easily separated from the seen variables. 

\subsubsection{Ablation Study}
As defined in Eq.(10), the overall objective of our model consists of $L_{f-SVAE}$, $L_{a-SVAE}$, $L_{cr}$ and $L_{cls}$. In this experiment, we analyze the impact of each term on AWA1 and CUB datasets. We report the AUC and the FPR scores on the threshold corresponding to $90\%$ TPR for four objective functions in Table \ref{tab:ablation}, where ``+'' stands for the combination of different terms. When there lacks of $L_{cr}$ and $L_{cls}$, we observe that the first objective function only achieves $81.5\%$ AUC score on AWA1, and $50.5\%$ on CUB. The FPR score are $51.9\%$ and $88.8\%$, respectively. It can be seen that the unseen samples can hardly be separated from the seen samples. When we further add $L_{cr}$, the AUC score increases to $83.0\%$ and the FPR decreases to $50.2\%$ on AWA1. The results also have improvements on CUB dataset. When we add $L_{cls}$ to the first objective function, the AUC score is improved to $85.3\%$ on AWA1 and $75.3\%$ on CUB. It can be seen that the classification loss heavily affects the binary classification. When we combine both $L_{cr}$ and $L_{cls}$, the overall objective achieves the best OOD classification results.    
\begin{table}[t]
    \scriptsize
	\renewcommand{\multirowsetup}{\centering}
	\renewcommand\arraystretch{1.0}
	\centering
	\caption{Binary classification results of different training objective functions. We report the AUC and the FPR corresponding to $90\%$ TPR. }
	\centering
	\resizebox{0.65\linewidth}{!}{
	\begin{tabular}{p{4cm}<{\centering}|
			p{0.75cm}<{\centering}p{0.75cm}<{\centering}|
			p{0.75cm}<{\centering}p{0.75cm}<{\centering}}
			
		\hline
		&\multicolumn{2}{c|}{AWA1}&	  \multicolumn{2}{c}{CUB}\\
		\hline
	     Objective Function  & \textbf{AUC}&	\textbf{FPR}&	
		         \textbf{AUC}&  \textbf{FPR}\\
		\hline
		 $L_{f-SVAE}+L_{a-SVAE}$ & 81.5 & 51.9 & 50.5 & 88.8  \\
		 $L_{f-SVAE}+L_{a-SVAE}+L_{cr}$ & 83.0 & 50.2 & 54.7 & 83.5\\
		 $L_{f-SVAE}+L_{a-SVAE}+L_{cls}$ & 85.3 & 47.4 & 75.3 & 71.5 \\
		 $L_{overall}$ & 85.9 & 45.2 & 75.9 & 69.8 \\
		\hline
	\end{tabular}}
	\label{tab:ablation}
\end{table}

\subsubsection{Parameters Sensitivity}
\begin{figure*}[t]
\begin{center}
    \includegraphics[width=0.5\linewidth]{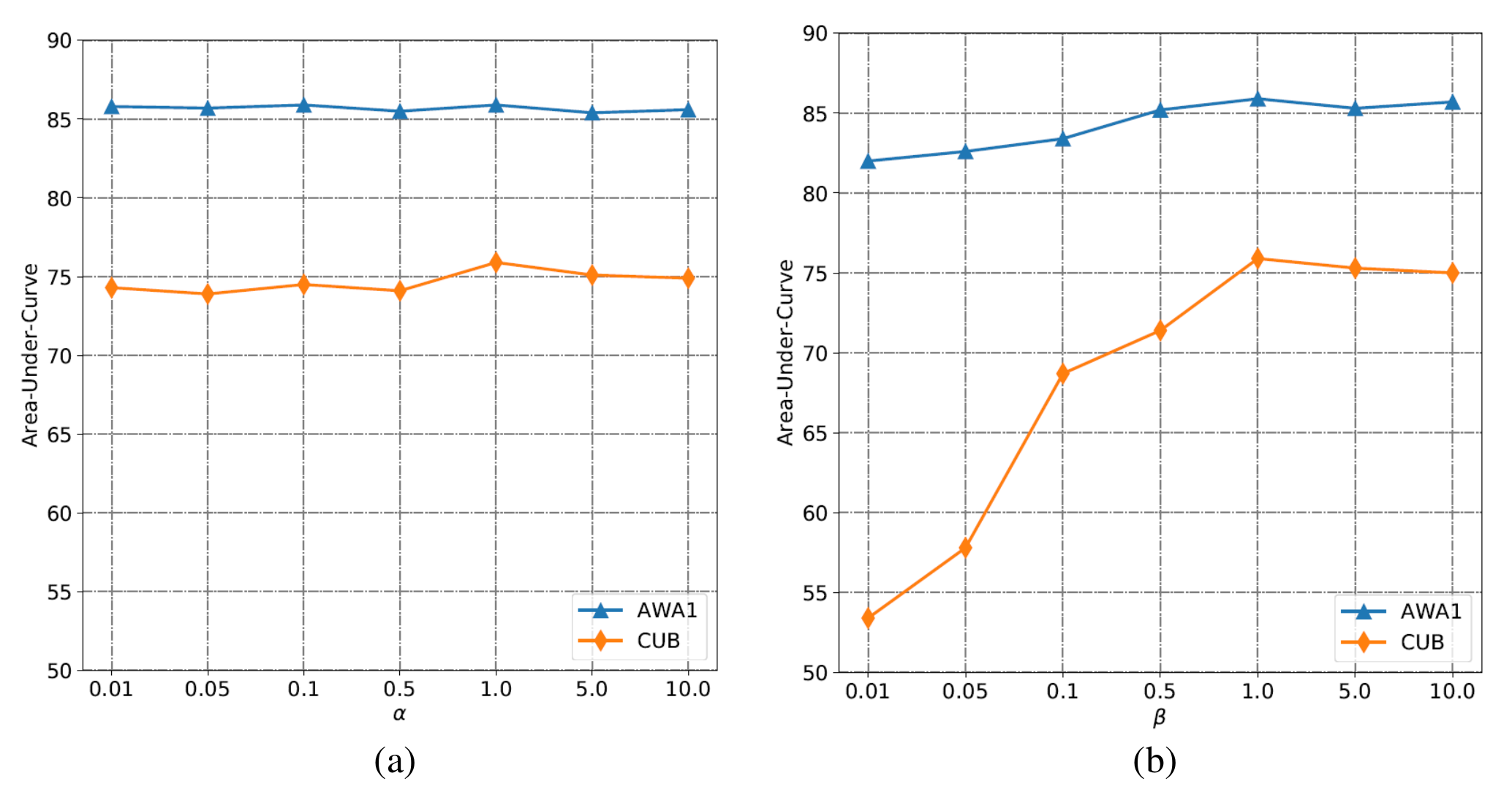}
\end{center}
   \caption{Parameter sensitivity on AWA1 and CUB datasets. }
\label{fig:hyper}
\end{figure*}
The hyper-parameters in our approach are tuned by cross-validation. Fixing $\lambda_f$ and $\lambda_a$  to 1.0, we mainly tune $\alpha$ and $\beta$ for our approach. Fig. \ref{fig:hyper} shows the AUC scores influenced by each hyper-parameter on AWA1 and CUB datasets, where each hyper-parameter is varied with the others are fixed. It can be seen that our method can work stably with different parameters.

\subsubsection{Dimension of Latent Space}
\begin{table}[h]
\caption{The influence of latent space dimension on the AUC score for AWA1 and CUB datasets.}
\center
\scriptsize
\renewcommand\arraystretch{1.0}
\resizebox{0.5\linewidth}{!}{
\begin{tabular}{c c c c c c c}
\hline
Dimension & $\quad $16$\quad $ &  $\quad $32$\quad $ &  $\quad $64$\quad $ &  $\quad $128$\quad $  &  $\quad $256$\quad $ \\ 
\hline
AWA1 & 85.3 & 85.6 & 85.5 & 85.9 & 85.4   \\
CUB & 71.0& 72.5 & 73.8  & 75.9 & 75.5  \\ 

\hline
\end{tabular}}

\label{tab:Latent_dimension}
\end{table}

In this analysis, we explore the robustness of our OOD classifier to the dimension of latent space. We report the AUC score in Table \ref{tab:Latent_dimension} with respect to different dimensions on AWA1 and CUB, ranging from 16, 32, 64, 128, and 256. We observe that the AUC score increases while we increase the latent space dimension and reaches the peak for both datasets at 128. When we continue to increase the dimension, the AUC score does not show obvious improvement. For general consideration, we set the dimension to 128 for all datasets.
\section{Conclusions}

In this paper, we present an Out-of-Distribution classifier for the Generalized Zero-Shot learning problem. The proposed classifier is based on multi-modal hyper-spherical variational auto-encoders which learns a bounded manifold for each seen class in the latent space. By using the boundaries, we can separate the unseen samples from the seen samples. After that, we use two experts to classify the unseen samples and the seen samples separately. In this way, the GZSL problem is simplified to a ZSL problem and a conventional supervised classification problem. We extensively evaluate our approach on five benchmark datasets. The experimental results demonstrate the advantages of our approach over state-of-the-art methods.

\section*{Acknowledgements}

This work was supported in part by Trico-Robot plan of NSFC under grant No.91748208, National Major Project under grant No.2018ZX01028-101, Shaanxi Project under grant No.2018ZDCXLGY0607, NSFC under grant No.61973246, and the program of the Ministry of Education for the university.

%
%
\bibliographystyle{splncs04}
\bibliography{egbib}
\end{document}